\documentclass[conference]{IEEEtran}
\IEEEoverridecommandlockouts

\usepackage{cite}
\usepackage{amsmath,amssymb,amsfonts}
\usepackage{algorithmic}
\usepackage{graphicx}
\usepackage{subcaption}
\usepackage{multirow}
\usepackage{booktabs}
\usepackage{textcomp}
\usepackage{xcolor}
\usepackage{url}
\usepackage{hyperref}

\def\BibTeX{{\rm B\kern-.05em{\sc i\kern-.025em b}\kern-.08em
    T\kern-.1667em\lower.7ex\hbox{E}\kern-.125emX}}
\begin{document}

\pagenumbering{arabic}
\pagestyle{plain}

\title{
Semi-Supervised Masked Autoencoders:
Unlocking Vision Transformer Potential with Limited Data


}


\author{
    \IEEEauthorblockN{Atik Faysal\textsuperscript{1},
    Mohammad Rostami\textsuperscript{1},
    Reihaneh Gh. Roshan\textsuperscript{2},
    Nikhil Muralidhar\textsuperscript{2} and
    Huaxia Wang\textsuperscript{1}}
    \IEEEauthorblockA{\textsuperscript{1}Department of Electrical and Computer Engineering, Rowan University, Glassboro, NJ, USA}
    \IEEEauthorblockA{\textsuperscript{2}Department of Computer Science, Stevens Institute of Technology, Hoboken, NJ, USA\\
    faysal24@rowan.edu, rostami23@rowan.edu, rghasemi@stevens.edu, nmurali1@stevens.edu, wanghu@rowan.edu}
}

\maketitle

\begin{abstract}

We address the challenge of training Vision Transformers (ViTs) when labeled data is scarce but unlabeled data is abundant. We propose Semi-Supervised Masked Autoencoder (SSMAE), a framework that jointly optimizes masked image reconstruction and classification using both unlabeled and labeled samples with dynamically selected pseudo-labels. SSMAE introduces a validation-driven gating mechanism that activates pseudo-labeling only after the model achieves reliable, high-confidence predictions that are consistent across both weakly and strongly augmented views of the same image, reducing confirmation bias. On CIFAR-10 and CIFAR-100, SSMAE consistently outperforms supervised ViT and fine-tuned MAE, with the largest gains in low-label regimes (+9.24\% over ViT on CIFAR-10 with 10\% labels). Our results demonstrate that when pseudo-labels are introduced is as important as how they are generated for data-efficient transformer training. Codes are available at \url{https://github.com/atik666/ssmae}.

    
\end{abstract}


\section{Introduction}

Vision Transformer (ViT) \cite{dosovitskiy2020image} has achieved remarkable success across various computer vision tasks, yet its performance heavily relies on large amounts of labeled data~\cite{steiner2021train}. In real-world scenarios, obtaining labeled data is often expensive, time-consuming, and requires domain expertise, while unlabeled data is abundant and readily available. This fundamental asymmetry has motivated extensive research in semi-supervised learning, which aims to leverage both limited labeled data and abundant unlabeled data to improve model performance \cite{touvron2021training}.

Recent advances in transformer-based semi-supervised learning have demonstrated the power of learning rich visual representations from unlabeled data \cite{weng2022semi, cai2022semi}. Masked Autoencoders (MAE)~\cite{he2022masked} have emerged as a particularly effective approach, learning to reconstruct masked portions of images through a Vision Transformer (ViT) encoder-decoder architecture. While MAE excels at learning generalizable features, it lacks the ability to directly incorporate labeled information for downstream classification tasks during pre-training.

To address this limitation, researchers have explored various strategies to integrate labeled data into the training process.  \cite{liang2022supmae} showed that inducing labelled data during unsupervised pretraining can help improve performance. Another promising direction in semi-supevised learning is the use of pseudo-labeling techniques, which generate artificial labels for unlabeled data based on the model's predictions \cite{yu2023inpl}. By leveraging high-confidence predictions, pseudo-labeling can effectively expand the labeled dataset and improve model performance. However, naive pseudo-labeling can suffer from confirmation bias, where incorrect predictions sabotage themselves, leading to degraded performance \cite{han2025regmixmatch}.

The key challenge in pseudo-labeling lies in determining when and how to generate reliable pseudo-labels. Existing approaches often rely on simple confidence thresholding~\cite{zheng2021rectifying} or complex adversarial training~\cite{pham2021meta}, but these methods may not adequately ensure the quality of generated pseudo-labels, especially in the early stages of training when model predictions are unreliable.

In this work, we propose SSMAE, a novel framework that combines the representational learning capabilities of MAE and a small amount of labeled data to give the classifier enough confidence for pseudo labeling. Addtionally, we introduce a confidence-based pseudo-labeling mechanism with several robust mechanism to ensure accurate pseudo label generation. Specifically, we generate pseudo-labels only when three conditions are simultaneously satisfied: (1) the model achieves high validation accuracy ($>70\%$) on sample it is highly confident ($>95\%$), (2) individual predictions exceed a high confidence threshold ($>95\%$) after a certain point of training; and (3) predictions remain consistent across different augmented views ( a weak and a strong augmented image) of the same image.

\begin{figure*}[!htbp]
    \centering
    \includegraphics[width=0.85\linewidth]{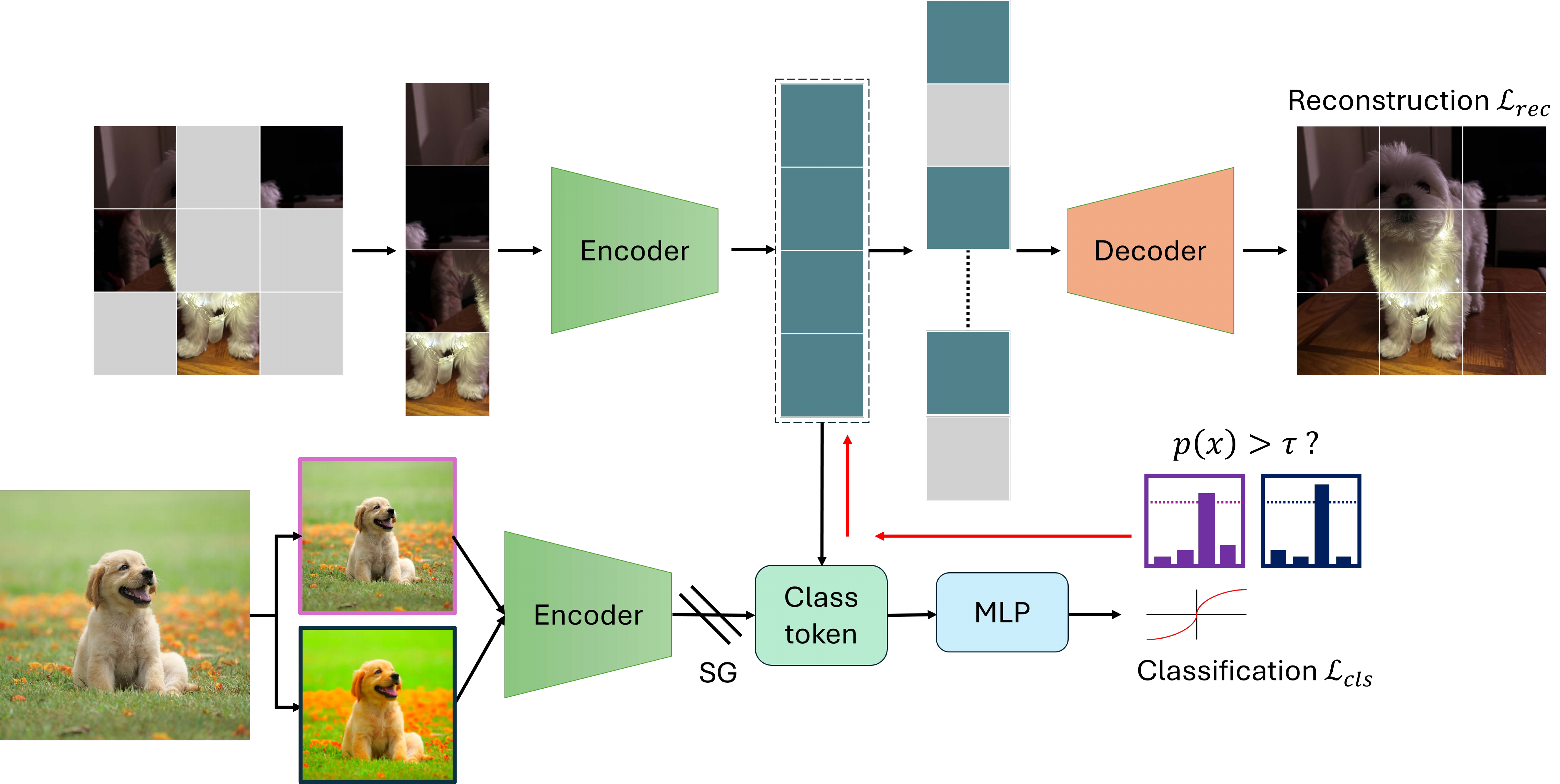}
    \caption{Overview of the SSMAE framework. A shared encoder is trained on two tasks: masked image reconstruction for all data, and classification for labeled data. For unlabeled data, our dynamic gate generates high-confidence pseudo-labels, which are then included in supervised classification.}
    \label{fig:ssmae_architecture}
\end{figure*}

The SSMAE architecture optimizes two complementary objectives: a self-supervised reconstruction loss that learns robust visual representations through masked image reconstruction, and a supervised classification loss that leverages both labeled data and high-confidence pseudo-labels. This dual-objective approach enables the model to benefit from both unsupervised feature learning and supervised classification signals. Our proposed framework is illustrated in Figure~\ref{fig:ssmae_architecture}.

Our main contributions are: 

\begin{enumerate}
    \item SSMAE combines unsupervised representation learning, supervised classification, and confidence-based pseudo-labeling, first masked model to achieve this integration.
    \item We propose a novel dynamic gating mechanism for pseudo-labeling that mitigates confirmation bias. This gate activates the use of pseudo-labels only after the model demonstrates reliability, measured by validation accuracy on high-confidence predictions and consistency across data augmentations.
    \item Since, the model learns through classification during pretraining, it does not require any further finetuning to achieve high accuracy on image classification tasks. However, finetuning can improve the accuracy further.
\end{enumerate}

\section{Related Work}
\label{sec:related}

\subsection{Semi-supervised Learning}
\label{subsec:ssl}

Semi-supervised learning aims to leverage both labeled and unlabeled data to improve model performance when labeled data is scarce. Earlier approaches include co-training~\cite{chen2022semi}, which trains multiple views of the data, and self-training methods that iteratively add confident predictions to the training set~\cite{shen2023co}. Recent advances have focused on consistency regularization techniques such as temporal ensembling~\cite{ke2019dual, zhang2021flexmatch} and mean teacher~\cite{xiong2022scmt, tarvainen2017mean}, which enforce consistent predictions under different augmentations of the same input.

More recently, contrastive learning methods like SimCLR~\cite{chen2020simple, chen2020big} and MoCo~\cite{he2020momentum} have shown promising results by learning representations that bring similar samples closer while pushing dissimilar ones apart.
Other methods like BYOL~\cite{grill2020bootstrap} and SimSiam~\cite{chen2021exploring} have further simplified this approach by demonstrating that strong representations can be learned without negative pairs. 

\subsection{Masking-based Representation Learning}
\label{subsec:masking}

Masking strategies have become fundamental in modern representation learning, popularized by BERT in natural language processing~\cite{devlin2019bert}. The core idea involves randomly masking portions of the input and training models to reconstruct the masked content, forcing the model to learn meaningful representations.

In computer vision, Masked Autoencoders (MAE)~\cite{he2022masked} demonstrated that this paradigm transfers effectively to images, achieving strong performance by masking large portions of image patches. Follow-up works have explored different masking strategies~\cite{li2022semmae}, multi-modal extensions~\cite{bachmann2022multimae}, and applications to video understanding~\cite{tong2022videomae}.

The success of masking-based methods stems from their ability to learn rich contextual representations without requiring labeled data, making them particularly suitable for self-supervised and semi-supervised learning scenarios.

\subsection{Pseudo-labeling}
\label{subsec:pseudo}

The key challenge in pseudo-labeling lies in selecting high-quality pseudo-labels while avoiding confirmation bias \cite{arazo2020pseudo}. Early approaches used simple confidence thresholding~\cite{hu2021simple}, but this can lead to overconfident predictions on out-of-distribution samples.

Recent advances have focused on improving pseudo-label quality through techniques such as consistency regularization~\cite{sohn2020fixmatch}, which combines confidence-based filtering with augmentation consistency, and uncertainty estimation~\cite{rizve2021defense}. Meta-learning approaches have also been proposed to learn better pseudo-labeling strategies~\cite{pham2021meta}.

Our work differs from existing pseudo-labeling methods by incorporating masking-based self-supervision to generate more robust pseudo-labels, particularly in scenarios with limited labeled data and domain shift.
\section{Methodology: SSMAE}

\subsection{Masking Strategy and Encoder Architecture}

Consider an input image $\mathbf{x} \in \mathbb{R}^{H \times W \times C}$. We first partition the image into 
\begin{equation}
N = \frac{HW}{P^2}
\label{eq:num_patches}
\end{equation}
non-overlapping patches of size $P \times P$. Let $\mathbf{x}_p \in \mathbb{R}^{N \times (P^2 C)}$ denote the flattened patches. A linear projection $E_{\text{patch}}$ maps these patches into a $d$-dimensional embedding space:
\begin{equation}
\mathbf{z}_p = E_{\text{patch}}(\mathbf{x}_p) \in \mathbb{R}^{N \times d}.
\label{eq:patch_embed}
\end{equation}

To preserve spatial relationships, we add learnable positional embeddings $\mathbf{p} \in \mathbb{R}^{N \times d}$, forming the initial token sequence:
\begin{equation}
\mathbf{Z}^{(0)} = \mathbf{z}_p + \mathbf{p}.
\label{eq:pos_embed}
\end{equation}

Following the MAE paradigm, we apply random masking to remove a fraction $r$ of tokens, retaining only the visible subset $\mathbf{Z}_v$. The Vision Transformer encoder $\mathcal{E}$ with $L$ layers processes only these visible tokens. Each layer $\ell$ applies multi-head self-attention (MSA) and feed-forward networks (FFN) with residual connections:
\begin{align}
\mathbf{Z}_v^{(\ell)} &= \mathbf{Z}_v^{(\ell-1)} + \text{MSA}\left(\text{LN}\left(\mathbf{Z}_v^{(\ell-1)}\right)\right), \label{eq:msa} \\
\mathbf{Z}_v^{(\ell)} &= \mathbf{Z}_v^{(\ell)} + \text{FFN}\left(\text{LN}\left(\mathbf{Z}_v^{(\ell)}\right)\right), \label{eq:ffn}
\end{align}
where LN denotes layer normalization. The encoder output provides latent representations:
\begin{equation}
\mathbf{h}_v = \mathcal{E}(\mathbf{Z}_v),
\label{eq:encoder_output}
\end{equation}
which serve dual purposes: feeding the decoder for masked patch reconstruction and providing the [CLS] token representation for classification tasks.

\subsection{Decoder Architecture and Reconstruction}

The decoder reconstructs the original image from the encoded visible tokens $\mathbf{h}_v$ and learnable mask tokens. We introduce shared mask token embeddings $\mathbf{m} \in \mathbb{R}^d$ for all masked positions, creating the full token sequence:
\begin{equation}
\mathbf{Z}_d^{(0)} = \text{Unshuffle}([\mathbf{h}_v \parallel \mathbf{M}]) + \mathbf{p}_d,
\label{eq:decoder_input}
\end{equation}
where $\mathbf{M} \in \mathbb{R}^{|\mathcal{I}_m| \times d}$ contains replicated mask tokens for masked indices $\mathcal{I}_m$, $\mathbf{p}_d \in \mathbb{R}^{N \times d}$ are decoder positional embeddings, and Unshuffle restores spatial order.

The decoder $\mathcal{D}$ applies $L'$ Transformer layers with the same architecture as the encoder but fewer parameters for efficiency. A linear projection maps the final decoder output to patch space:
\begin{equation}
\hat{\mathbf{x}} = \mathbf{Z}_d^{(L')} \mathbf{W}_{\text{out}} \in \mathbb{R}^{N \times (P^2 C)},
\label{eq:decoder_out}
\end{equation}
where $\mathbf{W}_{\text{out}} \in \mathbb{R}^{d \times (P^2 C)}$.

The reconstruction loss applies mean squared error only to masked patches:
\begin{equation}
\mathcal{L}_{\text{recon}} = \frac{1}{|\mathcal{I}_m|} \sum_{i \in \mathcal{I}_m} \|\hat{\mathbf{x}}_i - \mathbf{x}_i\|^2,
\label{eq:mae_loss}
\end{equation}
where $\hat{\mathbf{x}}_i$ and $\mathbf{x}_i$ are the reconstructed and original patches at masked position $i$.

\subsection{Classification Head}

For classification tasks, we prepend a learnable [CLS] token to the patch sequence. Importantly, during classification forward passes, we apply zero masking ratio ($r = 0$) to both labeled and unlabeled samples, ensuring all patches remain visible for feature extraction. The [CLS] token and all patch embeddings are processed by the encoder $\mathcal{E}$. We extract the [CLS] representation from the encoder output:
\begin{equation}
\mathbf{h}_{\text{cls}} = \mathbf{h}_v[0] \in \mathbb{R}^d,
\label{eq:cls_token}
\end{equation}

A linear classification head projects the [CLS] representation to class logits:
\begin{equation}
\mathbf{y} = \mathbf{h}_{\text{cls}} \mathbf{W}_{\text{cls}} + \mathbf{b}_{\text{cls}} \in \mathbb{R}^K,
\label{eq:classification}
\end{equation}
where $\mathbf{W}_{\text{cls}} \in \mathbb{R}^{d \times K}$ and $\mathbf{b}_{\text{cls}} \in \mathbb{R}^K$ are learnable parameters for $K$ classes.

For labeled samples $(\mathbf{x}_l, y_l)$, we compute the supervised classification loss:
\begin{equation}
\mathcal{L}_{\text{cls}}^{\text{sup}} = -\log p(y_l | \mathbf{x}_l),
\label{eq:sup_loss}
\end{equation}
where $p(y_l | \mathbf{x}_l) = \text{softmax}(\mathbf{y})_{y_l}$ is the predicted probability for the true class.

For unlabeled samples, we employ a confidence-based pseudo-labeling mechanism with consistency regularization. Given an unlabeled image $\mathbf{x}_u$, we apply weak augmentation $\mathcal{A}_w$ and strong augmentation $\mathcal{A}_s$, producing predictions:
\begin{align}
\mathbf{p}_w &= \text{softmax}(f(\mathcal{A}_w(\mathbf{x}_u))), \label{eq:weak_pred} \\
\mathbf{p}_s &= \text{softmax}(f(\mathcal{A}_s(\mathbf{x}_u))), \label{eq:strong_pred}
\end{align}
where $f(\cdot)$ denotes the complete model forward pass.

A pseudo-label $\hat{y}_u = \arg\max(\mathbf{p}_w)$ is accepted if:
\begin{align}
\max(\mathbf{p}_w) > \tau \text{ and } \max(\mathbf{p}_s) > \tau, \quad \text{(confidence)} \label{eq:confidence} \\
\arg\max(\mathbf{p}_w) = \arg\max(\mathbf{p}_s), \quad \text{(consistency)} \label{eq:consistency}
\end{align}
where $\tau$ is the confidence threshold. Accepted pseudo-labeled samples contribute to the classification loss:
\begin{equation}
\mathcal{L}_{\text{cls}}^{\text{pseudo}} = -\log p(\hat{y}_u | \mathcal{A}_s(\mathbf{x}_u)).
\label{eq:pseudo_loss}
\end{equation}

The total classification loss combines supervised and a weighted pseudo-supervised terms, $\lambda_{\text{p}}$, is given as follows:
\begin{equation}
\mathcal{L}_{\text{cls}} = \mathcal{L}_{\text{cls}}^{\text{sup}} +  \lambda_{\text{p}}  \cdot \mathcal{L}_{\text{cls}}^{\text{pseudo}}
\label{eq:total_cls_loss}
\end{equation}

\subsection{Dual-Mode Training Objective}

The training process optimizes a combined loss function that balances self-supervised and supervised objectives:

\begin{equation}
\mathcal{L}_{\text{total}} = \mathcal{L}_{\text{recon}} + \lambda \mathcal{L}_{\text{cls}}
\end{equation}

where $\mathcal{L}_{\text{recon}}$ is the masked autoencoder reconstruction loss computed on both labeled and unlabeled images, $\mathcal{L}_{\text{cls}}$ is the cross-entropy classification loss applied to labeled data and high-confidence pseudo-labeled samples and $\lambda$ is the classification loss weight.

\subsection{Dynamic Gating Mechanism}

To prevent noise from low-quality pseudo-labels, we activate pseudo-labeling only when the model is empirically reliable on confident predictions. The procedure is:

\textbf{Warm-up (no pseudo-labels)}: optimize $\mathcal{L}_{\text{recon}}$ on all images and $\mathcal{L}_{\text{cls}}^{\text{sup}}$ on labeled data for $T_{\text{warmup}}$ epochs (set $\lambda_{\text{p}}{=}0$).

\textbf{Reliability monitor}: after each epoch, evaluate a confidence-filtered validation accuracy on a held-out set $\mathcal{D}_{\text{val}}$ using the same confidence/consistency criteria as Eqs.~\eqref{eq:weak_pred}--\eqref{eq:consistency}.

\textbf{Gate activation}: let $g_t\in\{0,1\}$ denote the pseudo-label gate state depending on the confident validation accuracy, $\text{val}_{\text{conf}}^\text{{acc}}$ and threshold, $\tau_{\text{acc}}$, at epoch $t$.
    \begin{align*}
        g_t &= 1 \quad \text{activates if } \text{val}_{\text{conf}}^\text{{acc}} \ge \tau_{\text{acc}}, \\
        g_t &= 0 \quad \text{deactivates if } \text{val}_{\text{conf}}^\text{{acc}} < \tau_{\text{acc}}, 
    \end{align*}

\textbf{Loss integration}: include accepted samples in $\mathcal{L}_{\text{cls}}^{\text{pseudo}}$ (Eq.~\eqref{eq:pseudo_loss}). Set the effective weight to
    \[
        \lambda_{\text{p}}(t) = g_t \cdot \lambda_{\text{p}},
    \]

\textbf{Re-evaluation}: continue monitoring $\text{val}_{\text{conf}}^\text{{acc}}$. If it degrades below $\tau_{\text{acc}}$ for $n$ epochs, disable pseudo-labeling ($g_t{=}0$) to prevent error propagation.

\section{Experimental Setup}

\subsection{Datasets}

We evaluate our method on two widely used image classification datasets:

\textbf{CIFAR-10:} This dataset consists of 60,000 32×32 color images distributed across 10 classes. The standard split provides 50,000 images for training and 10,000 for testing.

\textbf{CIFAR-100:} Similar to CIFAR-10, this dataset contains 60,000 images but is significantly more challenging due to its 100 fine-grained classes.

For our semi-supervised experiments, we create subsets of the official training data, using only 10\%, 20\%, 30\%, and 40\% of the samples as labeled data. The remaining portion of the training set is treated as unlabeled data.

\subsection{Baselines}

To provide a comprehensive performance comparison, we evaluate SSMAE against two fundamental baselines:

\textbf{Supervised ViT:} A standard Vision Transformer (ViT-B/16) trained from scratch using only the available labeled subset of data. This baseline establishes the performance of a conventional supervised approach and highlights the challenges ViTs face in low-data regimes.

\textbf{MAE (Fine-tuned):} A standard Masked Autoencoder with a ViT-B/16 backbone, pre-trained in a self-supervised manner on the entire training set (both labeled and unlabeled portions, without using labels). Subsequently, the pre-trained encoder is fine-tuned on the available labeled subset.

\subsection{Implementation Details}

All the undertaken models use the ViT-Base (ViT-B/16) architecture. The encoder has an embedding dimension of 768, a depth of 12, and 12 attention heads. The MAE decoder is lighter, with an embedding dimension of 512, a depth of 8, and 16 attention heads.

The pre-training stage contains 200 epochs and fine-tuning stage has 100 epochs. We use the AdamW optimizer with a learning rate of $10^{-4}$ and a weight decay of $0.05$. We use a batch size of 16 for labeled data and 32 for unlabeled data. All experiments use an input image resolution of $224 \times 224$ pixels, with a patch size of $16 \times 16$. We use a high masking ratio of 75\%. 

For the dynamic gating mechanism, we set the warm-up period $T_{\text{warmup}}$ to 10 epochs, the high confidence validation accuracy threshold $\tau_{\text{acc}}$ is 70\%. The pseudo label confidence threshold $\tau$ is set to 0.95. The classification loss weight $\lambda$ is set to 1.0, and the pseudo-label loss weight $\lambda_{\text{p}}$ is set to 0.75. These particular parameters were empirically selected on a trial-and-validation basis.

\section{Results and Discussion}

\subsection{Reconstruction on Masked Patches}

\begin{figure*}[htbp]
    \centering
    \begin{subfigure}[b]{0.48\textwidth}
        \centering
        \includegraphics[width=\textwidth]{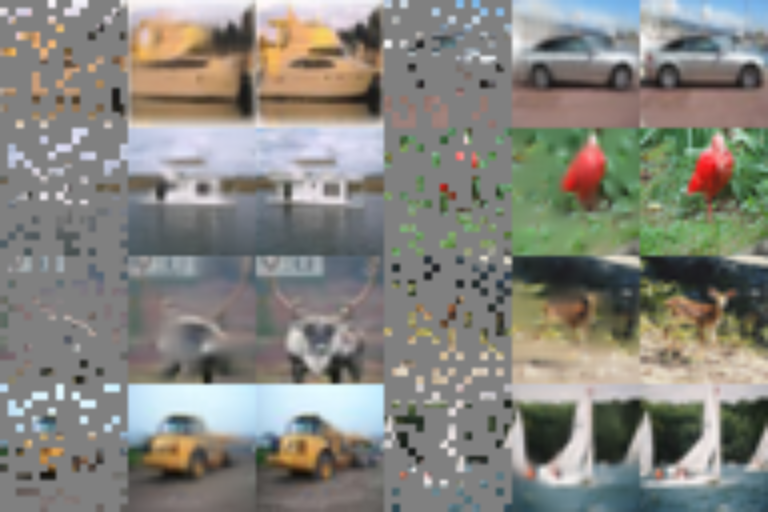}
        \caption{CIFAR-10 reconstruction results}
        \label{fig:cifar10_recon}
    \end{subfigure}
    \hfill
    \begin{subfigure}[b]{0.48\textwidth}
        \centering
        \includegraphics[width=\textwidth]{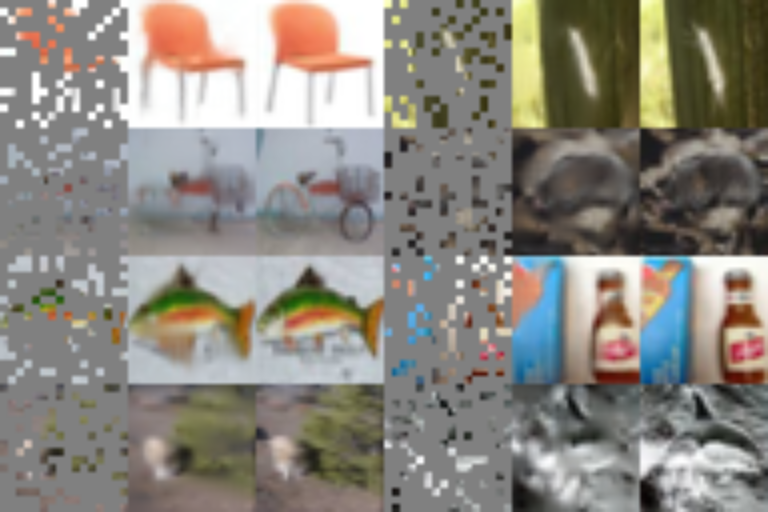}
        \caption{CIFAR-100 reconstruction results}
        \label{fig:cifar100_recon}
    \end{subfigure}
    \caption{Reconstruction results on CIFAR-10 and CIFAR-100 datasets. Each visualization shows three columns repreated once: masked input (left), SSMAE reconstruction (center), and original image (right).}
    \label{fig:reconstruction}
\end{figure*}

We first evaluate the reconstruction quality of SSMAE on masked patches of images from CIFAR-10 and CIFAR-100 datasets. Figure~\ref{fig:reconstruction} presents a comprehensive visualization of the reconstruction process, organized in three columns: the first column shows the masked input images with randomly excluded patches, the second column displays the corresponding reconstructions generated by our SSMAE model, and the third column presents the original unmasked images for comparison.

The results demonstrate that our SSMAE effectively reconstructs the missing patches across both datasets, successfully capturing fine-grained details and maintaining semantic consistency. On CIFAR-10, the model shows particularly strong performance in reconstructing natural objects and scenes, while on CIFAR-100's more diverse and challenging categories, the reconstructions preserve essential visual features despite the increased complexity. This reconstruction quality indicates that the model learns robust visual representations that generalize well across different levels of dataset complexity, even with limited labeled data in the semi-supervised setting.

\subsection{Classification Performance}

\textbf{CIFAR-100 Results:} SSMAE consistently outperforms baselines across all data regimes (Table~\ref{tab:cifar100_results}). In the most challenging 10\% labeled data setting, SSMAE achieves 22.65\% accuracy, representing a +0.93\% improvement over MAE (21.72\%) and +1.79\% over supervised ViT (20.86\%). Performance gains persist at higher data fractions, with SSMAE achieving 41.27\% at 40\% labeled data versus 40.08\% (MAE) and 39.10\% (ViT). The dual-objective training strategy demonstrates particular effectiveness in low-data regimes where traditional supervised approaches struggle.

\begin{table}[h]
\centering
\caption{Classification accuracy (\%) on CIFAR-100.}
\label{tab:cifar100_results}
\begin{tabular}{llllll}
\toprule
\multirow{3}{*}{\textbf{Models}} & \multirow{3}{*}{\textbf{Stage}} & \multicolumn{4}{c}{\textbf{Percentage of labeled data}} \\
\cmidrule(lr){3-6}
 & & \textbf{10\%} & \textbf{20\%} & \textbf{30\%} & \textbf{40\%} \\
\midrule
\multirow{2}{*}{SSMAE} & Pretrain & 19.65 & 28.57 & 33.18 & 35.96 \\
                       & Fine-tune & \textbf{22.65} & \textbf{32.41} & \textbf{35.31} & \textbf{41.27} \\
\midrule
MAE & Fine-tune & 21.72 & 30.20 & 34.88 & 40.08 \\
\midrule
ViT & Supervised & 20.86 & 28.28 & 34.04 & 39.10 \\
\bottomrule
\end{tabular}
\end{table}

\textbf{CIFAR-10 Results:} Similar trends emerge on CIFAR-10 (Table~\ref{tab:cifar10_results}), with SSMAE achieving substantial improvements in data-scarce scenarios. At 10\% labeled data, SSMAE reaches 56.80\% accuracy, significantly outperforming supervised ViT (+9.24\%) and MAE (+1.96\%). The confidence-based pseudo-labeling mechanism proves particularly effective on CIFAR-10's simpler class structure, with SSMAE maintaining competitive or superior performance across all data fractions tested.

The results validate our hypothesis that validation confidence serves as an effective gating mechanism for pseudo-label quality control. The consistent improvements across both datasets and data regimes demonstrate the robustness of the proposed semi-supervised framework in leveraging unlabeled data while maintaining representation learning quality through masked autoencoding.

\begin{table}[h]
\centering
\caption{Classification accuracy (\%) on CIFAR-10.}
\label{tab:cifar10_results}
\begin{tabular}{llllll}
\toprule
\multirow{3}{*}{\textbf{Models}} & \multirow{3}{*}{\textbf{Stage}} & \multicolumn{4}{c}{\textbf{Percentage of labeled data}} \\
\cmidrule(lr){3-6}
 & & \textbf{10\%} & \textbf{20\%} & \textbf{30\%} & \textbf{40\%} \\
\midrule
\multirow{2}{*}{SSMAE} & Pretrain & 54.13 & 61.48 & 63.49 & 70.13 \\
                       & Fine-tune & \textbf{56.80} & \textbf{66.40} & \textbf{71.83} & \textbf{74.67} \\
\midrule
MAE & Fine-tune & 54.84 &  64.58 & 71.07 & 72.79 \\
\midrule
ViT & Supervised & 47.56 & 64.05 & 69.55 & 72.07 \\
\bottomrule
\end{tabular}
\end{table}

\section{Ablation Studies}

\subsection{Ablation on Key Components}

\begin{table}[htbp]
\centering
\caption{Ablation on key components of SSMAE.}
\label{tab:ablation}
\begin{tabular}{lc}
\toprule
\textbf{Model Configuration} & \textbf{Accuracy (\%)} \\
\midrule
SSMAE (full model) & 66.40 \\
w/o recons. Loss ($\mathcal{L}_{cls}$ only with pseudo-labeling) & 64.17 \\
w/o consistency regularization (weak aug. only) & 61.77 \\
w/o dynamic gate (pseudo-labeling from epoch 1) & 62.37 \\
w/o dynamic gate (no validation accuracy threshold) & 63.49 \\
\bottomrule
\end{tabular}
\end{table}

To isolate the contributions of the key components within the SSMAE framework, we conducted an ablation study on CIFAR-10 using the 20\% labeled data setting. The results in Table~\ref{tab:ablation} clearly demonstrate that each component is critical to the framework's success. Removing the reconstruction loss and relying solely on classification with pseudo-labeling results in a performance drop from 66.40\% to 64.17\%. When consistency regularization is removed, accuracy falls significantly to 61.77\%. This underscores the vital role of enforcing consistent predictions between weakly and strongly augmented views of the same image, which is crucial for generating high-quality pseudo-labels.

We analyze our dynamic gating mechanism in two scenarios. First, removing the gate entirely and starting pseudo-labeling from the first epoch causes accuracy to drop to 62.37\%. This demonstrates the danger of confirmation bias from an immature model. Second, removing only the validation accuracy check results in an accuracy of 63.49\%. This shows that while a warm-up period is beneficial, the adaptive validation-based activation is key to maximizing performance by ensuring the model is sufficiently mature before incorporating pseudo-labels.

\subsection{Effect of Masking Ratio}

\begin{figure}[ht]
    \centering
    \includegraphics[width=0.95\linewidth]{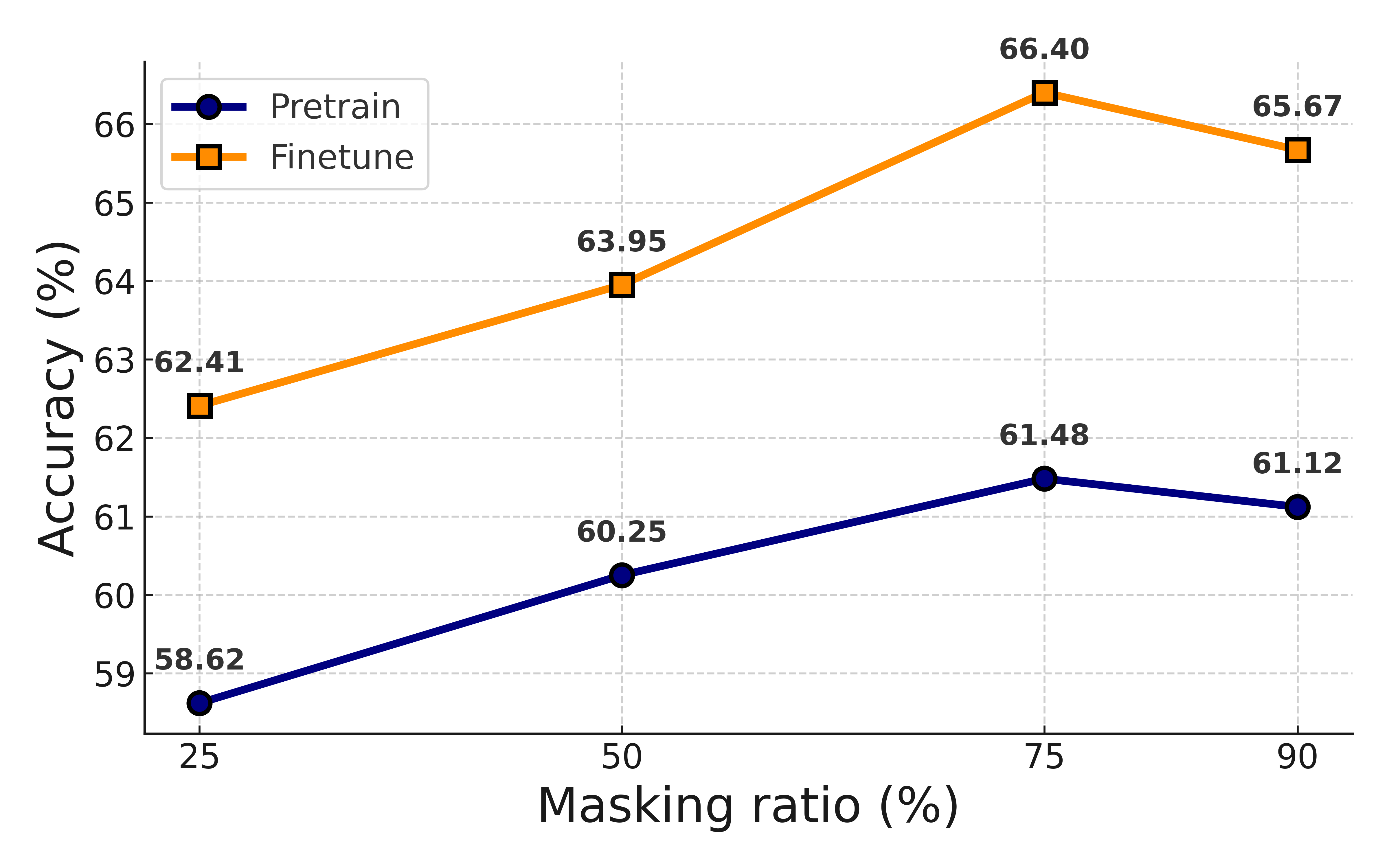}
    \caption{Effect of different masking ratios.}
    \label{fig:masking_ratio}
\end{figure}

We investigate the effect of the masking ratio on SSMAE's performance on CIFAR-10 with 20\% labeled data. As shown in Figure~\ref{fig:masking_ratio}, we find that a high masking ratio is beneficial. Performance peaks at a 75\% ratio, suggesting this provides a challenging self-supervision task that encourages learning robust features without removing too much context for reconstruction. A ratio of 90\% leads to a slight performance drop. Consequently, we use a 75\% masking ratio for all other experiments.

\section{Conclusion}

We presented SSMAE, a semi-supervised framework that unifies masked autoencoding and dynamic pseudo-labeling to enable data-efficient ViT training. Central to SSMAE is a validation-driven gating mechanism that defers pseudo-label usage until the model demonstrates reliable, high-confidence predictions, effectively mitigating early-stage confirmation bias. Experiments on CIFAR-10 and CIFAR-100 show consistent gains over both supervised ViT and fine-tuned MAE, particularly in low-label regimes. Ablations confirm that the reconstruction objective, consistency regularization, and dynamic gating each contribute substantially to performance. Our results highlight that controlling when pseudo-labels are introduced is as critical as how they are generated, offering a simple yet powerful strategy for scaling large models under scarce supervision.

\bibliographystyle{IEEEtran}
\bibliography{references}

@inproceedings{he2022masked,
  title={Masked autoencoders are scalable vision learners},
  author={He, Kaiming and Chen, Xinlei and Xie, Saining and Li, Yanghao and Doll{\'a}r, Piotr and Girshick, Ross},
  booktitle={Proceedings of the IEEE/CVF conference on computer vision and pattern recognition},
  pages={16000--16009},
  year={2022}
}

@article{dosovitskiy2020image,
  title={An image is worth 16x16 words: Transformers for image recognition at scale},
  author={Dosovitskiy, Alexey and Beyer, Lucas and Kolesnikov, Alexander and Weissenborn, Dirk and Zhai, Xiaohua and Unterthiner, Thomas and Dehghani, Mostafa and Minderer, Matthias and Heigold, Georg and Gelly, Sylvain and others},
  journal={arXiv preprint arXiv:2010.11929},
  year={2020}
}

@article{steiner2021train,
  title={How to train your vit? data, augmentation, and regularization in vision transformers},
  author={Steiner, Andreas and Kolesnikov, Alexander and Zhai, Xiaohua and Wightman, Ross and Uszkoreit, Jakob and Beyer, Lucas},
  journal={arXiv preprint arXiv:2106.10270},
  year={2021}
}

@inproceedings{weng2022semi,
  title={Semi-supervised vision transformers},
  author={Weng, Zejia and Yang, Xitong and Li, Ang and Wu, Zuxuan and Jiang, Yu-Gang},
  booktitle={European conference on computer vision},
  pages={605--620},
  year={2022},
  organization={Springer}
}

@article{cai2022semi,
  title={Semi-supervised vision transformers at scale},
  author={Cai, Zhaowei and Ravichandran, Avinash and Favaro, Paolo and Wang, Manchen and Modolo, Davide and Bhotika, Rahul and Tu, Zhuowen and Soatto, Stefano},
  journal={Advances in Neural Information Processing Systems},
  volume={35},
  pages={25697--25710},
  year={2022}
}

@inproceedings{touvron2021training,
  title={Training data-efficient image transformers \& distillation through attention},
  author={Touvron, Hugo and Cord, Matthieu and Douze, Matthijs and Massa, Francisco and Sablayrolles, Alexandre and J{\'e}gou, Herv{\'e}},
  booktitle={International conference on machine learning},
  pages={10347--10357},
  year={2021},
  organization={PMLR}
}

@article{liang2022supmae,
  title={Supmae: Supervised masked autoencoders are efficient vision learners},
  author={Liang, Feng and Li, Yangguang and Marculescu, Diana},
  journal={arXiv preprint arXiv:2205.14540},
  year={2022}
}

@article{yu2023inpl,
  title={Inpl: Pseudo-labeling the inliers first for imbalanced semi-supervised learning},
  author={Yu, Zhuoran and Li, Yin and Lee, Yong Jae},
  journal={arXiv preprint arXiv:2303.07269},
  year={2023}
}

@inproceedings{han2025regmixmatch,
  title={RegMixMatch: Optimizing Mixup Utilization in Semi-Supervised Learning},
  author={Han, Haorong and Yuan, Jidong and Wei, Chixuan and Yu, Zhongyang},
  booktitle={Proceedings of the AAAI Conference on Artificial Intelligence},
  volume={39},
  number={16},
  pages={17032--17040},
  year={2025}
}

@article{zheng2021rectifying,
  title={Rectifying pseudo label learning via uncertainty estimation for domain adaptive semantic segmentation},
  author={Zheng, Zhedong and Yang, Yi},
  journal={International Journal of Computer Vision},
  volume={129},
  number={4},
  pages={1106--1120},
  year={2021},
  publisher={Springer}
}

@inproceedings{pham2021meta,
  title={Meta pseudo labels},
  author={Pham, Hieu and Dai, Zihang and Xie, Qizhe and Le, Quoc V},
  booktitle={Proceedings of the IEEE/CVF conference on computer vision and pattern recognition},
  pages={11557--11568},
  year={2021}
}

@inproceedings{chen2022semi,
  title={Semi-supervised learning with multi-head co-training},
  author={Chen, Mingcai and Du, Yuntao and Zhang, Yi and Qian, Shuwei and Wang, Chongjun},
  booktitle={Proceedings of the AAAI conference on artificial intelligence},
  volume={36},
  number={6},
  pages={6278--6286},
  year={2022}
}

@article{shen2023co,
  title={Co-training with high-confidence pseudo labels for semi-supervised medical image segmentation},
  author={Shen, Zhiqiang and Cao, Peng and Yang, Hua and Liu, Xiaoli and Yang, Jinzhu and Zaiane, Osmar R},
  journal={arXiv preprint arXiv:2301.04465},
  year={2023}
}

@inproceedings{ke2019dual,
  title={Dual student: Breaking the limits of the teacher in semi-supervised learning},
  author={Ke, Zhanghan and Wang, Daoye and Yan, Qiong and Ren, Jimmy and Lau, Rynson WH},
  booktitle={Proceedings of the IEEE/CVF international conference on computer vision},
  pages={6728--6736},
  year={2019}
}

@inproceedings{xiong2022scmt,
  title={SCMT: Self-Correction Mean Teacher for Semi-supervised Object Detection.},
  author={Xiong, Feng and Tian, Jiayi and Hao, Zhihui and He, Yulin and Ren, Xiaofeng},
  booktitle={IJCAI},
  pages={1488--1494},
  year={2022}
}

@article{zhang2021flexmatch,
  title={Flexmatch: Boosting semi-supervised learning with curriculum pseudo labeling},
  author={Zhang, Bowen and Wang, Yidong and Hou, Wenxin and Wu, Hao and Wang, Jindong and Okumura, Manabu and Shinozaki, Takahiro},
  journal={Advances in neural information processing systems},
  volume={34},
  pages={18408--18419},
  year={2021}
}

@article{tarvainen2017mean,
  title={Mean teachers are better role models: Weight-averaged consistency targets improve semi-supervised deep learning results},
  author={Tarvainen, Antti and Valpola, Harri},
  journal={Advances in neural information processing systems},
  volume={30},
  year={2017}
}

@inproceedings{chen2020simple,
  title={A simple framework for contrastive learning of visual representations},
  author={Chen, Ting and Kornblith, Simon and Norouzi, Mohammad and Hinton, Geoffrey},
  booktitle={International conference on machine learning},
  pages={1597--1607},
  year={2020},
  organization={PmLR}
}

@article{chen2020big,
  title={Big self-supervised models are strong semi-supervised learners},
  author={Chen, Ting and Kornblith, Simon and Swersky, Kevin and Norouzi, Mohammad and Hinton, Geoffrey E},
  journal={Advances in neural information processing systems},
  volume={33},
  pages={22243--22255},
  year={2020}
}

@inproceedings{he2020momentum,
  title={Momentum contrast for unsupervised visual representation learning},
  author={He, Kaiming and Fan, Haoqi and Wu, Yuxin and Xie, Saining and Girshick, Ross},
  booktitle={Proceedings of the IEEE/CVF conference on computer vision and pattern recognition},
  pages={9729--9738},
  year={2020}
}

@article{grill2020bootstrap,
  title={Bootstrap your own latent-a new approach to self-supervised learning},
  author={Grill, Jean-Bastien and Strub, Florian and Altch{\'e}, Florent and Tallec, Corentin and Richemond, Pierre and Buchatskaya, Elena and Doersch, Carl and Avila Pires, Bernardo and Guo, Zhaohan and Gheshlaghi Azar, Mohammad and others},
  journal={Advances in neural information processing systems},
  volume={33},
  pages={21271--21284},
  year={2020}
}

@inproceedings{chen2021exploring,
  title={Exploring simple siamese representation learning},
  author={Chen, Xinlei and He, Kaiming},
  booktitle={Proceedings of the IEEE/CVF conference on computer vision and pattern recognition},
  pages={15750--15758},
  year={2021}
}

@inproceedings{devlin2019bert,
  title={Bert: Pre-training of deep bidirectional transformers for language understanding},
  author={Devlin, Jacob and Chang, Ming-Wei and Lee, Kenton and Toutanova, Kristina},
  booktitle={Proceedings of the 2019 conference of the North American chapter of the association for computational linguistics: human language technologies, volume 1 (long and short papers)},
  pages={4171--4186},
  year={2019}
}

@article{li2022semmae,
  title={Semmae: Semantic-guided masking for learning masked autoencoders},
  author={Li, Gang and Zheng, Heliang and Liu, Daqing and Wang, Chaoyue and Su, Bing and Zheng, Changwen},
  journal={Advances in Neural Information Processing Systems},
  volume={35},
  pages={14290--14302},
  year={2022}
}

@inproceedings{bachmann2022multimae,
  title={Multimae: Multi-modal multi-task masked autoencoders},
  author={Bachmann, Roman and Mizrahi, David and Atanov, Andrei and Zamir, Amir},
  booktitle={European Conference on Computer Vision},
  pages={348--367},
  year={2022},
  organization={Springer}
}

@article{tong2022videomae,
  title={Videomae: Masked autoencoders are data-efficient learners for self-supervised video pre-training},
  author={Tong, Zhan and Song, Yibing and Wang, Jue and Wang, Limin},
  journal={Advances in neural information processing systems},
  volume={35},
  pages={10078--10093},
  year={2022}
}

@inproceedings{arazo2020pseudo,
  title={Pseudo-labeling and confirmation bias in deep semi-supervised learning},
  author={Arazo, Eric and Ortego, Diego and Albert, Paul and O’Connor, Noel E and McGuinness, Kevin},
  booktitle={2020 International joint conference on neural networks (IJCNN)},
  pages={1--8},
  year={2020},
  organization={IEEE}
}

@inproceedings{hu2021simple,
  title={Simple: Similar pseudo label exploitation for semi-supervised classification},
  author={Hu, Zijian and Yang, Zhengyu and Hu, Xuefeng and Nevatia, Ram},
  booktitle={Proceedings of the IEEE/CVF conference on computer vision and pattern recognition},
  pages={15099--15108},
  year={2021}
}

@article{sohn2020fixmatch,
  title={Fixmatch: Simplifying semi-supervised learning with consistency and confidence},
  author={Sohn, Kihyuk and Berthelot, David and Carlini, Nicholas and Zhang, Zizhao and Zhang, Han and Raffel, Colin A and Cubuk, Ekin Dogus and Kurakin, Alexey and Li, Chun-Liang},
  journal={Advances in neural information processing systems},
  volume={33},
  pages={596--608},
  year={2020}
}

@article{rizve2021defense,
  title={In defense of pseudo-labeling: An uncertainty-aware pseudo-label selection framework for semi-supervised learning},
  author={Rizve, Mamshad Nayeem and Duarte, Kevin and Rawat, Yogesh S and Shah, Mubarak},
  journal={arXiv preprint arXiv:2101.06329},
  year={2021}
}

\end{document}